\documentclass[11pt,a4paper]{article}
\usepackage{authblk}
\usepackage[hyperref]{acl2021}
\usepackage{times}
\usepackage{latexsym}
\usepackage{graphicx}
\usepackage{booktabs}
\usepackage{linguex}

\usepackage{microtype}

\aclfinalcopy 

\title{LIIR at SemEval-2021 task 6: Detection of Persuasion Techniques In Texts and Images using CLIP features}

\author[ ]{Erfan Ghadery}
\author[ ]{Damien Sileo}
\author[ ]{Marie-Francine Moens}

\affil[ ]{Department of Computer Science (CS)}
\affil[ ]{
KU Leuven}
\affil[ ]{{\{erfan.ghadery, damien.sileo, sien.moens\}@kuleuven.be}}

\date{}

\begin{document}
\maketitle
\begin{abstract}
We describe our approach for SemEval-2021 task 6 on detection of persuasion techniques in multimodal content (memes). Our system combines pretrained multimodal models (CLIP) and chained classifiers. Also, we propose to enrich the data by a data augmentation technique. Our submission achieves a rank of 8/16 in terms of F1-micro and 9/16 with F1-macro on the test set.
\end{abstract}

\section{Introduction}

Online propaganda is potentially harmful to society, and the task of automated propaganda detection has been suggested to alleviate its risks \citep{ijcai2020-672}.
In particular, providing a justification when performing propaganda detection is important for acceptability and application of the decisions.
Previous challenges have focused on the detection of propaganda techniques \citep{martino2020SemEval2020}, based on news articles.
However, many use cases do not solely involve text, but can also involve other modalities, notably images. Task 6 of SemEval-2021 proposes a shared task on the detection of persuasion techniques detection in memes, where both images and text are involved. Substasks 1 and 2 deal with text in isolation, but we focus on subtask 3: visuolinguistic persuasion technique detection. 
\par This article presents the system behind our submission for subtask 3 \cite{semeval21}. To handle this problem, we use a modelinproceedings{guerini-etal-2008-resources containing three components: data augmentation, image and text feature extraction, and chain classifier components. First, given a paired image-text as the input, we paraphrase the text part using back-translation and pair it again with the corresponding image to enrich the data. Then, we extract visual and textual features using the CLIP \cite{radford2021learning} image encoder and text encoder, respectively. Finally, we use a chain classifier to model the relation between labels for the final prediction. Our proposed method, named LIIR, has achieved a competitive performance with the best performing methods in the competition. Also, empirical results show that the augmentation approach is effective in improving the results.

\par The rest of the article is organized as follows. The next section reviews related works. Section 3 describes the methodology of our proposed method. We will discuss experiments and evaluation results in Sections 4 and 5, respectively.  Finally, the last section contains the conclusion of our work.

\section{Related work}
\begin{center}
\begin{figure*}[t]
    \centering
    \includegraphics[clip, trim=0cm 2.5cm 1cm 0cm, width=1\textwidth]{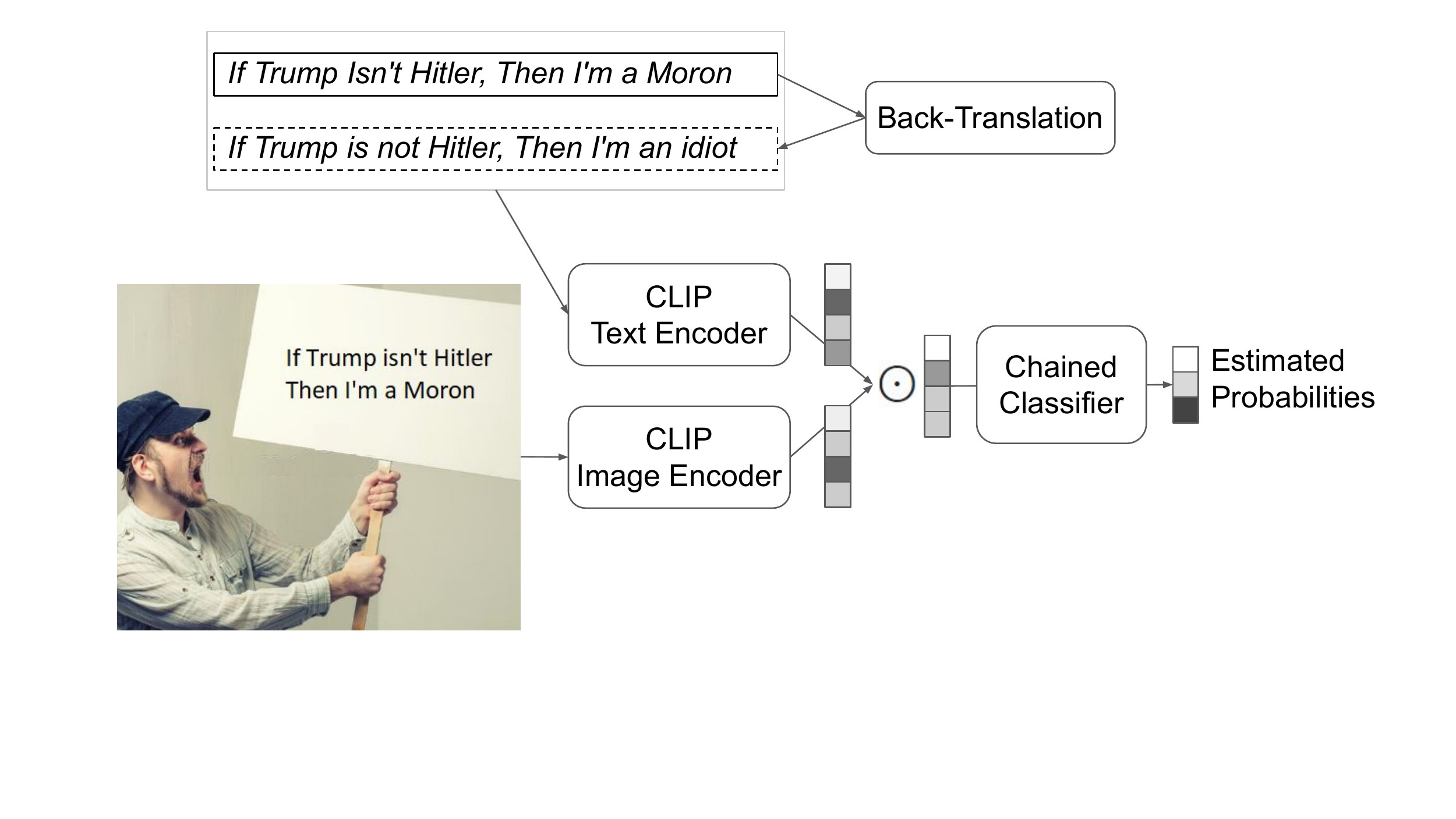}
    \caption{The overall architecture of our proposed model. For each example, use Back-Translation to derive augmentations of the text, and we compute persuasion techniques probabilities separately. Then, we average the estimated probabilities from augmented and original examples.}
    \label{diagram}
\end{figure*}
\end{center}

This work is related to computational techniques for automated propaganda detection \citep{ijcai2020-672} and is the continuation of a previous shared task \citep{martino2020SemEval2020}.

Taks 11 of SemEval-2020 proposes a more fine-graind analysis by also identifying the underlying techniques behind propaganda in news text, with annotations derived from previously proposed propaganda techniques typologies \cite{miller1939detect,weston2000rulebook}.

This current iteration of the task tackles a more challenging domain, by including multimodal content, notably memes. The subtle interaction between text and image is an open challenge for state of the art multimodal models. For instance, the Hateful Memes challenge \citep{kiela2020hateful} was recently proposed, as a binary task for detection of hateful content. The recent advances in pretraining of visuolinguistic representations \citep{chen2020uniter} lead the model closer to human accuracy \citep{sandulescu2020detecting}.

More generally, propaganda detection is at the crossroad of many tasks, since it can be helped by many subtasks. Fact-checking \cite{Aho:72, dale2017nlp} can be involved with propaganda detection, alongside various social, emotional and discursive aspects \cite{ DBLP:journals/corr/abs-1907-08672}, including offensive language detection \cite{Pradan2020, ghadery2020liir} emotion analysis \citep{dolan2002emotion}, computational study of persuasiveness \citep{guerini-etal-2008-resources,Persuasion2018Ng} and argumentation \citep{PalauMoens2009, habernal2016makes}.

\section{Methodology}
In this section, we introduce the design of our proposed method. The overall architecture of our method is depicted in figure \ref{diagram}. Our model consists of several components: a data augmentation component (Back-translation), a feature extraction component(CLIP), and a chained classifier. Details of each component are described in the following subsections.

\subsection{Augmentation Method}
One of the challenges in this subtask is the low number of training data where the organizers have provided just 200 training samples. To enrich the training set we propose to use the back-translation technique \cite{sennrich2015improving} for paraphrasing a given sentence by translating it to a specific target language and translating back to the original language. To this end, we use four translation models, English-to-German, German-to-English, English-to-Russian, and Russian-to-English provided by \cite{ng2019facebook}. Therefore, for each training sentence, we obtain two paraphrased version of it. In the test time, we average the probability distributions over the original and paraphrased sentence-image pairs.

\subsection{Feature Extraction}

Our system isProbabilities  of a combination of pretrained  visuolinguistic and linguistic models.

We use CLIP \citep{radford2021learning} as a pretrained visuolinguistic model. CLIP provides an image encoder $f_i$ and a text encoder $f_t$. They were pretrained on a prediction of matching image/text pairs. The training objective incentivizes high values of $f_i(I).f_t(T)$ if $I$ and $T$ are matching in the training corpus, and low values of they are not matching\footnote{They assign each image to the text associated to other images in the current batch to generate negative examples}.
Instead of using a dot product, we create features with element-wise product $f_i(I)\odot f_t(T)$ of image and text encoding. This enables aspect-based representations of the matching between image and text. We experimented with other compositions \cite{sileo-etal-2019-composition} which did not lead to significant improvement.

We then use a classifier $C$ on top of $f_i(I)\odot f_t(T)$ to predict the labels.

\subsection{Chained Classifier}

In this task, we are dealing with a multilabel classification problem, which means we need to predict a subset of labels for a given paired image-text sample as the input. We noticed that label co-occurrences were not uniformly distributed, as shown in figure \ref{fig:co-oc}. To further address the data sparsity, we use another inductive bias at the classifier-level with a chained classifier \cite{read2009classifier} using scikit-learn implementation \citep{scikit-learn}.

Instead of considering each classification task independently, a chained classifier begins with the training of one classifier for each of the $L$ labels. But we also sequentially train $L$ other classifier instances thereafter, each of them using the outputs of the previous classifier as input. This allows our model to model the correlations between labels. We use a Logistic Regression with default parameter as our base classifier.

\par Our chain classifier uses combined image and text features as the input. We transfer the predicted probabilities of the classifier via the sigmoid activation function to make the probability values more discriminating\cite{ghadery2018unsupervised}. 
Then we apply thresholding on the $L$ labels probabilities since the task requires a discrete set of labels as output.
We predict a label when the associated probability is above a given threshold. We optimize the threshold on the validation set by a simple grid search using values between 0.0 and 0.9 with a step of 0.005.

\begin{figure}
  \centering
\includegraphics[width=0.48\textwidth]{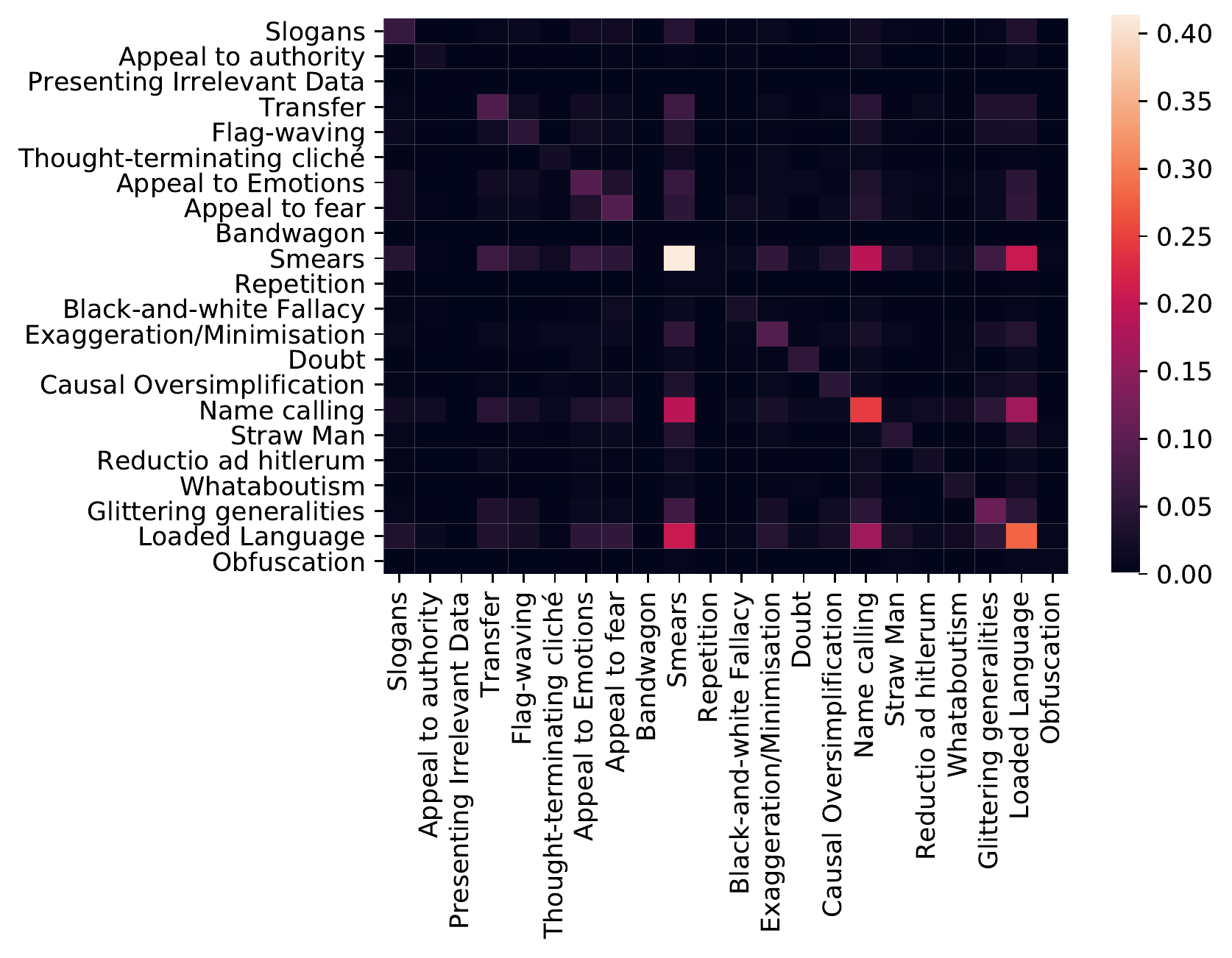}
  \caption{Probabilities of label co-occurence in the training set. Some label pairs, for instance (\textsc{Smears} and \textsc{Loaded Language}) are frequently associated. \label{fig:co-oc}}
\end{figure}

\section{Experiments}
\subsection{Datasets}
We use the dataset provided by SemEval-2021 organizers for task 6. The dataset consists of 687(290) samples as the training set, 63 samples as the dev set, and 200 samples as the test set. Each sample is an image and its corresponding text. We use 10$\%$ of the training set as the validation set for hyperparameter tuning.

\begin{table}
\footnotesize
\centering

\label{tab:labels}
\begin{tabular}{lr}
\toprule
                                             Label &  Count \\
\midrule
                                            Smears &    199 \\
                                   Loaded Language &    134 \\
                             Name calling/Labeling &    118 \\
                  Glittering generalities (Virtue) &     54 \\
                       Appeal to (Strong) Emotions &     43 \\
                          Appeal to fear/prejudice &     42 \\
                         Exaggeration/Minimisation &     42 \\
                                          Transfer &     41 \\
                                           Slogans &     28 \\
                                             Doubt &     25 \\
                                       Flag-waving &     24 \\
                         Causal Oversimplification &     22 \\
 Misrepresentation of Someone's Position  &     21 \\
                                      Whataboutism &     14 \\
              Black-and-white Fallacy/Dictatorship &     13 \\
                        Thought-terminating cliché &     10 \\
                              Reductio ad hitlerum &     10 \\
                               Appeal to authority &     10 \\
                                        Repetition &      3 \\
     Obfuscation, Intentional vagueness, Confusion &      3 \\
                                         Bandwagon &      1 \\
          Presenting Irrelevant Data (Red Herring) &      1 \\
\bottomrule
\end{tabular}
\caption{Labels of persuasion techniques with associated counts in the training set}
\end{table}

\section{Evaluation and Results}
\subsection{Results}
In this section, we present the results obtained by our model on the test sets for Subtask 3. Table \ref{res_test} shows the results obtained by the submitted final model on the test set. All the results are provided in terms of macro-F1 and Micro-F1. Furthermore, we provide the results obtained by the random baseline, the best performing method in the competition, and median result for the sake of comparison. Note that, we used the first released training set at the time of final submission which contained just 290 training samples. Therefore, we also provide results obtained by our model after using all the provided 687 training samples. Results show that LIIR has achieved a good performance compared to the majority class baseline and the median result which demonstrates that our model can effectively identify persuasion techniques in text and images. Also, we can observe LIIR has achieved a competitive performance compared to the best result obtained by the best team in the competition when it uses all the training samples.

\begin{table}[h!]
\centering
\begin{tabular}{|l|c|c|}
\hline
\textbf{System}              & \textbf{Macro-F1} & \textbf{Micro-F1} \\ \hline
Majority class &     0.05152   &  0.07062        \\ \hline
Median             & 0.18842 & 0.4896          \\ \hline
LIIR(290 examples)             & 0.18807 & 0.49835         \\ \hline
LIIR(687 examples)             & 0.21796 & 0.51122         \\ \hline
Best system             & \textbf{0.27315} & \textbf{0.58109}          \\ \hline

\end{tabular}
\caption{The results obtained by LIIR compared to the baselines on the Test set for Subtask 3. Numbers in parentheses show the total number of train samples used by our model.}
\label{res_test}
\end{table}

\subsection{Ablation Analysis}
In this part, we provide an ablation study on the effect of different components of our proposed method on the dev set. First, we show the effect of using just visual features, just textual features, and both. Furthermore, we examine how well the final results of our model was influenced by the augmentation method. Table \ref{ablation1} shows the ablation study on the effect of using different features. The first observation is that image features contain more information compared to the textual features. Also, we can observe that the best Micro-F1 score is obtained when we combine both visual and textual features. These results show the effectiveness of our method in making use of both visual and textual information.

\begin{table}[h!]
\centering
\small
\begin{tabular}{|l|c|c|}
\hline
\textbf{System}              & \textbf{Macro-F1} & \textbf{Micro-F1} \\ \hline
LIIR \hspace{0.3cm} -- textual features &     0.32275   &  0.53237        \\ \hline
LIIR \hspace{0.3cm} -- visual features &     \textbf{0.33347}   &  0.52954        \\ \hline
LIIR             & 0.29972 & \textbf{0.58312}          \\ \hline
\end{tabular}
\caption{Ablation analysis for the effect of using different features by our model on the dev set.}
\label{ablation1}
\end{table}

In Table \ref{ablation2}, the effect of the augmentation technique is shown. As the results show, the augmentation approach is quite effective in improving the model performance by a high margin.

\begin{table}[h!]
\centering
\small
\begin{tabular}{|l|c|c|}
\hline
\textbf{System}              & \textbf{Macro-F1} & \textbf{Micro-F1} \\ \hline
LIIR \hspace{0.3cm} w/o Augmentation &     0.25090   &  0.54952        \\ \hline

LIIR \hspace{0.3cm} w Augmentation            & \textbf{0.29972} & \textbf{0.58312}          \\ \hline
\end{tabular}
\caption{Ablation analysis for the effect of augmentation method on the dev set.}
\label{ablation2}
\end{table}

\section{Negative Results}
We also tried to use CLIP as a zero-shot classifier for propaganda technique detection.
To do so, we constructed prompts such as :

\ex. \textit{This image is committing \textsc{[LABEL]} fallacy.} 

or

\ex. \textit{Saying that \textsc{[TEXT]} is  \textsc{[LABEL] } fallacy.}

For each input image/text, we generated a prompt for each labels, and used CLIP to estimate the estimate an affinity score between the prompt and the image. CLIP is designed to predict relatedness between the input image and text, and we expected that an input text mentioning the relevant propaganda technique should be associated with higher probabilities that the others.

However, this method  did not seem to perform better than chance. This suggests that propaganda detection technique task might be too abstract for CLIP in zero-shot settings.

\section{Conclusion}
We described our submission for the shared task of multimodal propaganda technique detection at SemEval-2021. Our system performances that are competitive with other systems even though we used a simple architecture with no ensemble, by leveraging non-supervised learning. We believe that further work on zero-shot learning would be a valuable way to improve propaganda detection techniques for the least frequent labels.

\section{Acknowledgments}
This research was funded by the CELSA project from the KU Leuven with grant number CELSA/19/018.
\bibliographystyle{acl_natbib}
\bibliography{main}


\end{document}